\documentclass[runningheads]{llncs}

\usepackage{eccv}

\usepackage{eccvabbrv}

\usepackage{graphicx}
\usepackage{booktabs}
\usepackage{multirow}
\usepackage{bm}
\usepackage{wrapfig}
\newcommand{\bhline}[1]{\noalign{\hrule height #1}}

\usepackage[accsupp]{axessibility}  

\usepackage{hyperref}

\usepackage{orcidlink}

\begin{document}

\title{
Generalized SAM: Efficient Fine-Tuning of SAM for Variable Input Image Sizes
} 

\titlerunning{Generalized SAM}

\author{Sota Kato\inst{1}\orcidlink{0000-0003-0392-6426} \and
Hinako Mitsuoka\inst{1}\orcidlink{0009-0005-6969-4017} \and
Kazuhiro Hotta\inst{1}\orcidlink{0000-0002-5675-8713}}

\authorrunning{S.~Kato et al.}

\institute{
Meijo University, 1-501 Shiogamaguchi, Tempaku-ku, Nagoya, 468-8502, Japan\\
\email{150442030@ccalumni.meijo-u.ac.jp}\\
\email{200442165@ccalumni.meijo-u.ac.jp}\\
\email{kazuhotta@meijo-u.ac.jp}
}

\maketitle

\begin{abstract}
There has been a lot of recent research on improving the efficiency of fine-tuning foundation models.
In this paper, we propose a novel efficient fine-tuning method that allows 
the input image size of Segment Anything Model (SAM) to be variable. 
SAM is a powerful foundational model for image segmentation trained on huge datasets, but it requires fine-tuning to recognize arbitrary classes.
The input image size of SAM is fixed at $1024 \times 1024$, resulting in substantial computational demands during training. 
Furthermore, the fixed input image size may result in the loss of image information, \eg due to fixed aspect ratios.
To address this problem, we propose Generalized SAM (GSAM).
Different from the previous methods, GSAM is the first to apply random cropping during training with SAM, thereby significantly reducing the computational cost of training. 
Experiments on datasets of various types and various pixel counts have shown that GSAM can train more efficiently than SAM and other fine-tuning methods for SAM, achieving comparable or higher accuracy.
Our code will be available at: 
\textcolor{magenta}{
\url{https://github.com/usagisukisuki/G-SAM}}.
\end{abstract}

\begin{figure*}[t]
\begin{center}
\includegraphics[scale=0.33]{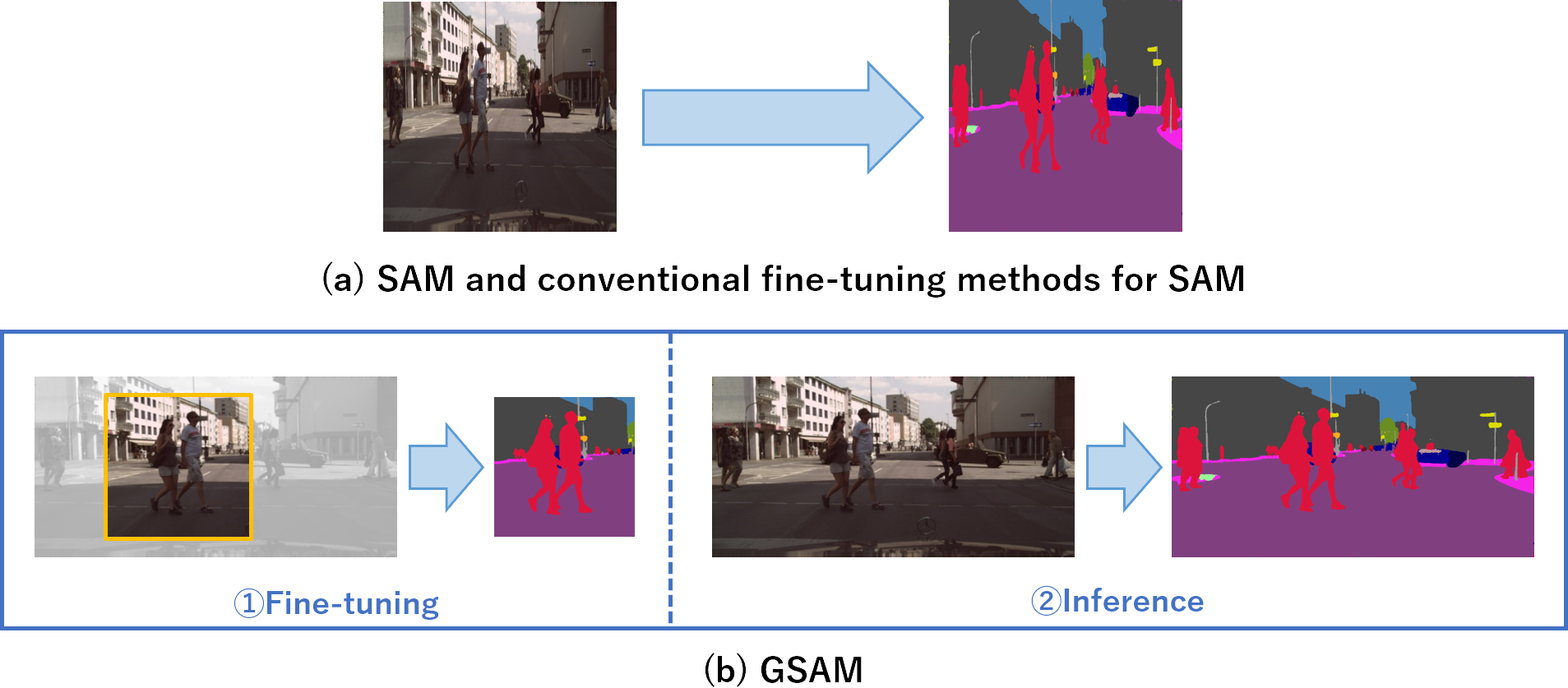}
\end{center}
\caption{When SAM is fine-tuned for semantic segmentation with conventional methods, only fixed-size images can be input. As a result, input images are deformed to fit a specific size, causing information loss. In contrast, GSAM supports various input image sizes while maintaining the superior segmentation performance of SAM. This allows images to be used in their original form and enables random cropping during fine-tuning, previously unavailable in SAM-related methods. GSAM provides efficient fine-tuning, specialized for semantic segmentation of arbitrary data, minimizing information loss and computational costs.}
\label{fig:intro}
\end{figure*}

\begin{figure*}[t]
\begin{center}
\includegraphics[scale=0.45]{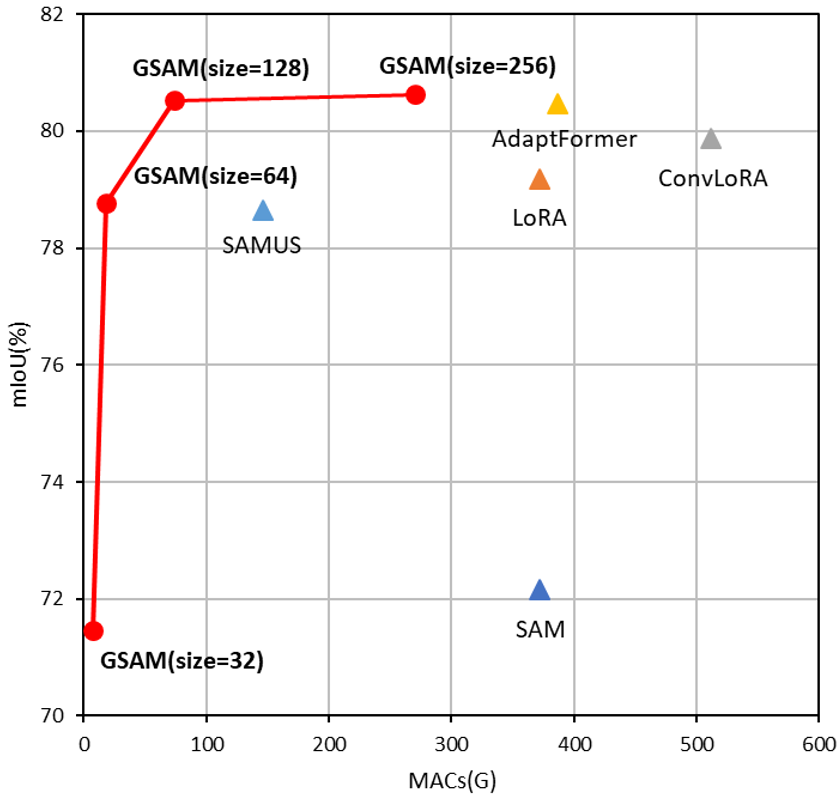}
\end{center}
\caption{The \textbf{trade-off} between MACs and segmentation accuracy (mIoU) for conventional fine-tuning methods for SAM on the ISBI2012 dataset \cite{isbi}. The Red circles indicate our proposed GSAM and the triangles indicate the conventional methods. Random cropping is only performed for GSAM, cropping to the number of pixels indicated by the "size". Random cropping cannot be used except for GSAM due to its structure.}
\label{fig:tradeoff}
\end{figure*}

\section{Introduction}
\label{sec:intro}
Deep learning has been widely applied to various image recognition problems with great success \cite{imagenet, resnet, vit}.
In particular, in recent years, a large-scale and comprehensive model called the foundation model, has been proposed, and it is known to be a powerful model that can achieve high performance for a wide range of tasks \cite{bert, clip, gpt4}.
In the field of semantic segmentation, SAM \cite{sam} was proposed in 2023 and can perform highly accurate segmentation on natural images without training.
However, if we want to identify arbitrary classes using SAM, we need to perform fine-tuning using the teacher labels of the target dataset. 
Since the input image size for SAM is fixed at $1024 \times 1024$, this causes a huge computational cost problem during fine-tuning.

Although methods such as LoRA \cite{lora} and AdaptFormer \cite{adaptformer} have been proposed for fine-tuning SAM more effectively, the input image size for these methods is fixed to $1024 \times 1024$ is the same as for SAM, and the problem of computational cost due to the input image size has not been solved.
A fine-tuning method for SAM that reduces the input image size to SAM and can train on small images such as $256 \times 256$ has also been proposed \cite{samus}, but again the input image size must be fixed. 
As the number of pixel counts varies in each dataset, the use of a fixed number of pixel counts is likely to lead to serious problems such as missing image information.

In this paper, we propose \textbf{G}eneralized \textbf{SAM} (GSAM), which can train even when the input image size is variable. 
In the conventional segmentation models based on Convolutional Neural Networks (CNN) \cite{unet, deeplabv3plus} which were proposed before SAM, segmentation was possible even if the input image size during training and inference were different, so it is possible to input a small random cropped image during training and input the original image size during inference to obtain segmentation results.
As shown in \cref{fig:intro}, GSAM is the first method using SAM that can apply random cropping at training time, and the use of a small random cropping size reduces the computational cost at training. 
The fixed input size of SAM is due to fixed-size Positional Encoding.
Therefore, GSAM supports variable input image sizes by employing a Positional Encoding Generator (PEG) \cite{peg} consisting of a Depth-wise Convolution layer as a substitute for Positional Encoding.
Furthermore, we also propose Spatial-Multiscale (SM) AdaptFormer to consider more spatial information during fine-tuning. 
SM-AdaptFormer has a multi-scale structure and can handle feature vectors integrating a more diverse and wider range of spatial information. 
This is a segmentation-specific fine-tuning method since proper segmentation requires information at various scales.

From the evaluation experiments on seven different datasets consisting of in-vehicle images, satellite images, microscopic images, endoscopic images, CT images, and transparent object images, we confirmed that the proposed GSAM can significantly reduce the computational cost of training compared to the conventional fine-tuning methods for SAM, and achieved comparable or higher segmentation accuracy.
As shown in \cref{fig:tradeoff}, GSAM achieved the trade-off of lower computational cost and higher accuracy by enabling random cropping.
In particular, on the Synapse multi-organ dataset, which is CT images, GSAM achieved segmentation accuracy of more than 11\% better than conventional SAM fine-tuning methods, indicating that our proposed method may be highly effective in certain areas.

This paper is organized as follows.
\cref{sec:relatedwork} describes the related works.
\cref{sec:method} describes the details of our proposed method.
\cref{sec:experiments} shows the experimental results. 
Finally, the conclusion and future works are described in \cref{sec:conclusion}. 

Our \textbf{contributions} can be summarized as follows:
\begin{quote}
    \begin{itemize}
        \item We propose a novel efficient fine-tuning method for SAM, GSAM. 
        GSAM can cope with variable input image sizes, allowing random cropping to be used the first time during fine-tuning for SAM.
        \item We also propose SM-AdaptFormer to acquire multi-scale features during fine-tuning for SAM.
        \item From the evaluation experiments on various datasets, we confirmed that GSAM can significantly reduce the computational cost of training compared to the conventional SAM fine-tuning methods, and achieved comparable or higher segmentation accuracy.     
    \end{itemize}
\end{quote}

\section{Related Works}
\label{sec:relatedwork}

\subsection{Segmentation Models}
\label{sub:semseg}

Since U-Net \cite{unet} revolutionized the area of semantic segmentation of images, various architectures have been proposed to improve accuracy \cite{unet++,r2unet}.
Methods such as PSPNet \cite{pspnet} and DeepLab series \cite{deeplab, deeplabv3, deeplabv3plus}, which specialize in obtaining features at various scales, and more recently, methods based on Transformer, have also emerged \cite{transunet, swinunet}.
Compared to these methods, GSAM does not require a particularly complex structure and only requires efficient fine-tuning of the foundation model, SAM, to adapt to semantic segmentation to achieve competitive performance.

\subsection{Foundation Models}
\label{sub:foundation}

Since Transformer \cite{transformer} was published in 2017, various foundational models have been built due to its amazing extensibility.
Foundation models such as BERT \cite{bert}, LLaMa \cite{llama}, and GPT-4 \cite{gpt4} have shown ground-breaking performance in natural language processing.
Recently, there has been a remarkable development of foundation models in the field of computer vision, with many high-performance models such as Segment Anything Model (SAM) \cite{sam}, CLIP \cite{clip}, and Stable Diffusion \cite{sd}.
Among others, SAM is a segmentation model trained on huge datasets with high zero-shot generalization performance.
However, foundation models generally have high generalization performance but lack expertise and require fine-tuning to recognize specific downstream tasks, and arbitrary classes properly.
For this reason, a lot of research is being done to effectively and efficiently fine-tune foundation models such as SAM.

\subsection{Efficient Fine-tuning for SAM}
\label{sub:effient}

When we fine-tune foundation models with a huge number of parameters such as SAM, it is computationally very expensive to update all parameters.
It is therefore common to update only a part of the weight parameters to achieve fine-tuning at a lower computational cost.
Low-Rank Adaptation (LoRA) \cite{lora} successfully reduces the number of learnable parameters in downstream tasks by applying learnable low-rank matrices to each Transformer layer. 
This method originated in the field of natural language processing, but it has also been adapted to computer vision and can be used to fine-tune SAM.
ConvLoRA \cite{convlora}, which applies convolutional layers to LoRA and reinforces image-related local priors to achieve higher accuracy, has also been proposed.
Additionally, AdaptFormer \cite{adaptformer} achieves higher accuracy with minimal additional learnable parameters by using two fully connected layers and an activation function in each Feed-Forward Network (FNN).

However, the input image size for these methods is fixed at $1024 \times 1024$ which is the same as SAM, and thus the computational cost issue related to the input image size has not been resolved. 
In this paper, we propose to reduce the computational cost of fine-tuning SAM by using smaller input images only during training.

\subsection{Changing The Input Image Size for SAM}
\label{sub:changepixel}

Recently, some methods have been proposed that allow training with smaller images by reducing the input image size of SAM from $1024 \times 1024$.
SAMed \cite{samed} enables the input image size of $512 \times 512$ by applying LoRA to SAM. 
Additionally, SAMUS \cite{samus} achieves high accuracy in medical image segmentation even with a smaller input image size of $256\times 256$ by integrating the feature maps of Transformer and CNN using Cross Attention.

However, if the images to be handled are larger than the input size, there is a possibility that image information may be lost due to resizing.
In this case, a method that enables segmentation even if the size of the input images differs between training and inference is needed.

\section{Proposed Method}
\label{sec:method}

\begin{figure*}[t]
\begin{center}
\includegraphics[scale=0.425]{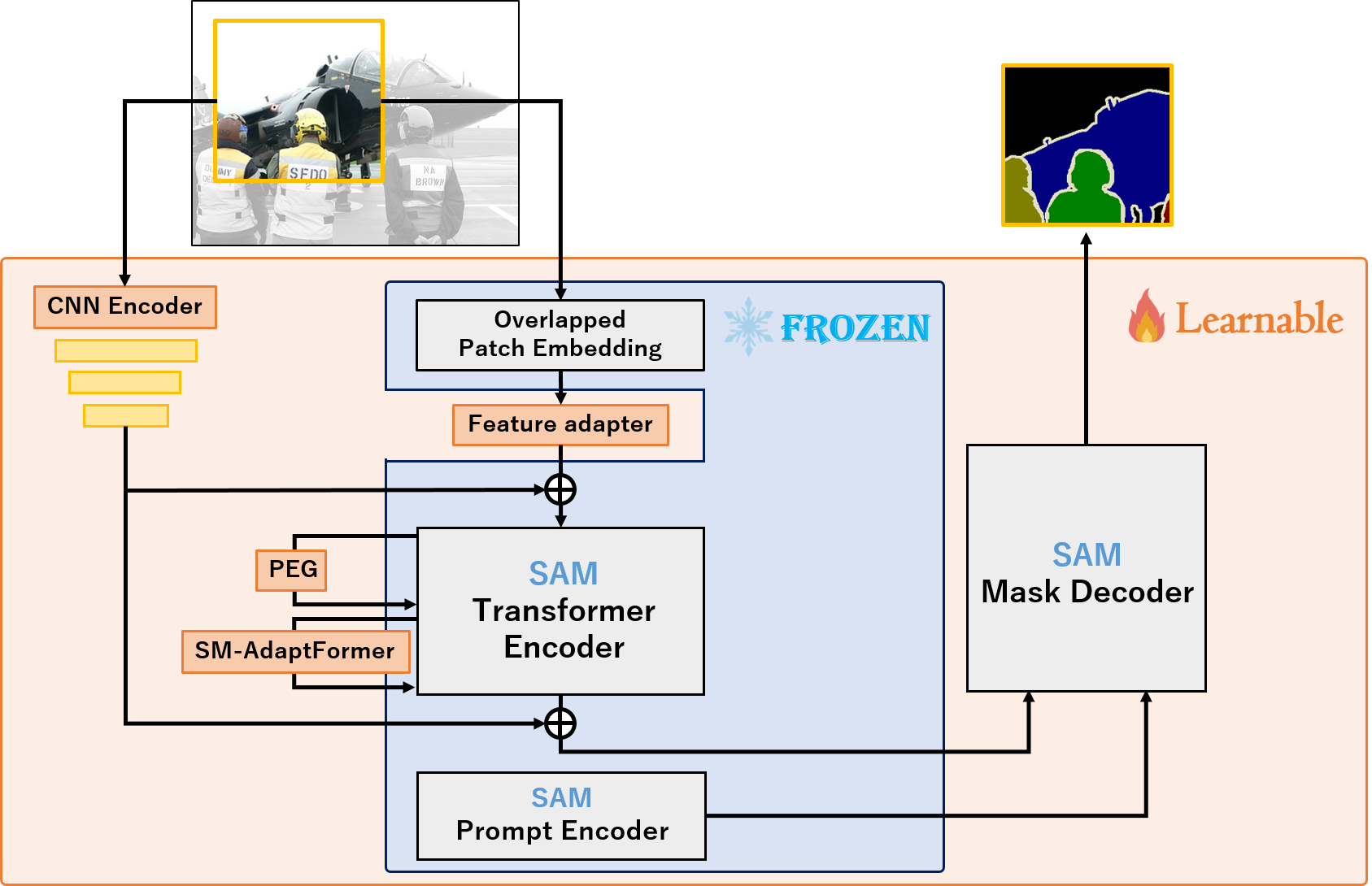}
\end{center}
\caption{Overview of Generalized SAM. $FROZEN$ indicates a network in which the weight parameters are fixed, and $Learnable$ indicates a network in which the weight parameters are updated.}
\label{fig:gsam}
\end{figure*}

In this paper, in order to efficiently fine-tune with random cropping in training, we propose a novel method called Generalized SAM (GSAM).
\cref{fig:gsam} illustrates the overview of GSAM. 
In GSAM, all weight parameters of the Prompt Encoder and some weight parameters of the Transformer Encoder are fixed, and the other weight parameters are updated during fine-tuning. 
In addition, GSAM adds a novel structure to use random cropping in training. 
The details of each structure of GSAM are described \cref{sub:randcrop}, \cref{sub:smadapt}.

\subsection{Application of Random Cropping during Training}
\label{sub:randcrop}

Since the input to SAM must be of a fixed size of $1024 \times 1024$, it is impossible to handle random cropping with small image sizes during training.
The most important reason why the input to SAM must be fixed is that the Positional Encoding in the Transformer Encoder, which is a component of SAM, is of a fixed size. 
Positional Encoding is a structure that adds information to each token to inform its own position of the Vision Transformer.
In the case of SAM, it is a learnable weight parameter with a fixed size.
Therefore, GSAM employs the Positional Encoding Generator (PEG) \cite{peg} as a substitute for Positional Encoding.
PEG consists of a Depth-wise Convolution layer that considers only spatial orientation, which enables it to retain positional information even when the input size of the feature map is variable.

However, the original pretrained SAM does not support random inputs, and it is possible that global learning by Self-Attention in the Transformer Encoder alone is insufficient for small and variable inputs. 
Therefore, we use a new network composed of CNNs,
shown in \cref{fig:gsam} as CNN Encoder in order to learn by integrating CNN features and SAM features.
Since learning by local kernels of CNN is effective for smaller input images, it is considered to complement the feature map of the Transformer Encoder in SAM.
GSAM adds the feature maps of the third block of ResNet101 \cite{resnet}, which retains some spatial information, to pre-input and post-output feature maps of the Transformer Encoder.
This enables efficient fine-tuning using random cropping. 
From the above, the introduction of PEG and CNN Encoder enables the use of random cropping and corresponding feature extraction.

\begin{figure*}[t]
\begin{center}
\includegraphics[scale=0.42]{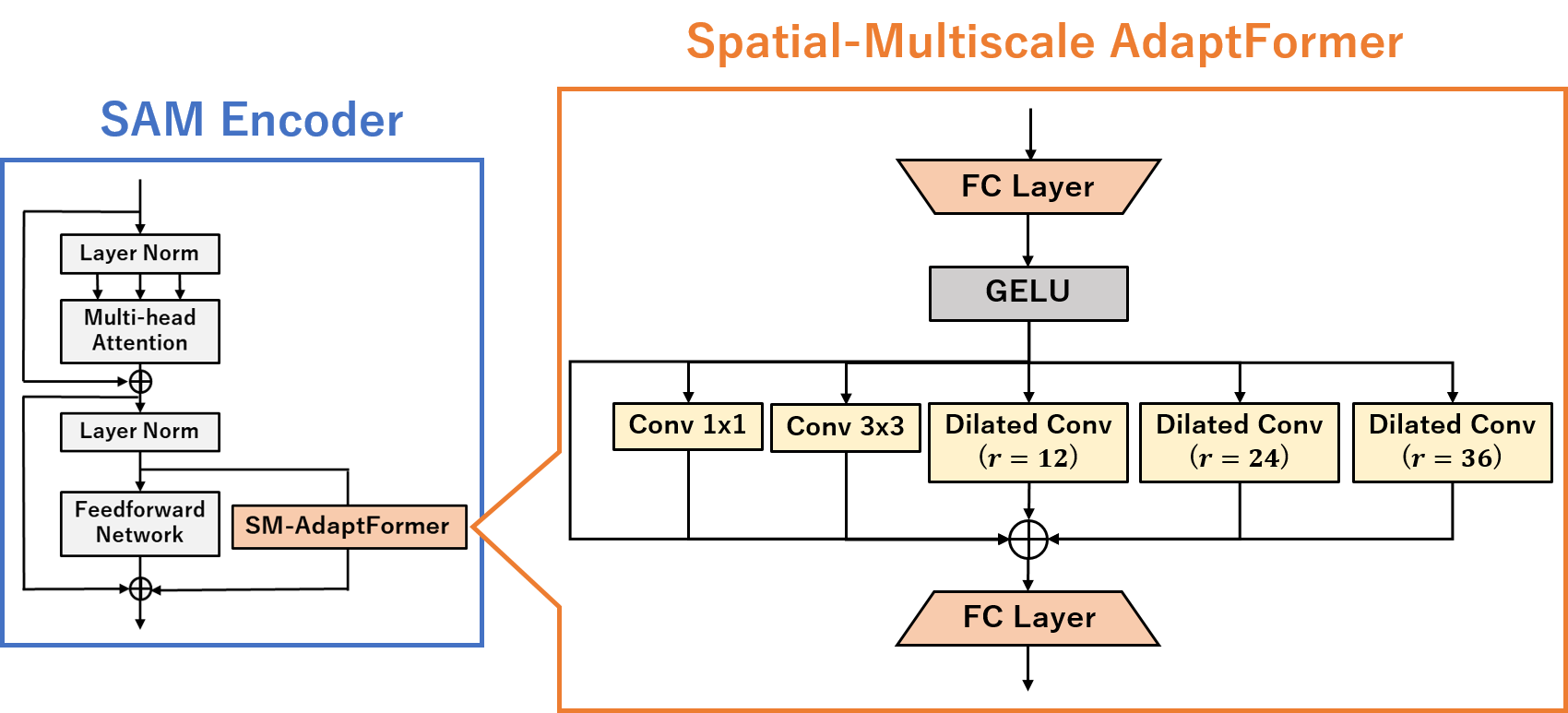}
\end{center}
\caption{Overview of Spatial-Multiscale AdaptFormer. Five convolutional layers with different receptive fields are used to acquire the spatial features necessary for semantic segmentation.}
\label{fig:sm}
\end{figure*}

\subsection{Spatial-Multiscale AdaptFormer}
\label{sub:smadapt}

In order to further improve the discrimination accuracy for the target dataset, we propose Spatial-Multiscale (SM) AdaptFormer. 
\cref{fig:sm} illustrates the overview of SM-AdaptFormer. 
AdaptFormer \cite{adaptformer} is known as a low computational cost and high performance method for fine-tuning SAM, but AdaptFormer does not take spatial information into account. 
Since spatial features are important information in semantic segmentation, the proposed SM-AdaptFormer prepares multiple convolutional layers with kernels in various ranges and acquires multiscale features.

For convolutional layers with a wide range of kernels, we employ Dilated Convolution \cite{dilatedconv}. 
Dilated Convolution can expand the receptive field while maintaining the same kernel size, allowing global feature extraction without increasing computational cost.
When Dilated Convolution is applied to the input feature map $\bm{x}$, the output feature map $\bm{y}$ can be expressed as in \cref{eq:dilation} using the position $\bm{i}$ of each pixel and the convolution kernel $\bm{w}$. 
\begin{equation}
    \bm{y\lbrack i\rbrack} = \sum_k \bm{x\lbrack i}+r \cdot \bm{k\rbrack w\lbrack k\rbrack}.
    \label{eq:dilation}
\end{equation}
where $r$ is a parameter that determines the width of the stride, and the receptive field of the convolutional layer can be adaptively changed by changing $r$. 
SM-AdaptFormer provides two convolutional layers with kernel sizes of $1 \times 1$ and $3 \times 3$, as well as Dilated Convolution ($r=12$), Dilated Convolution ($r=24$), and Dilated Convolution ($r=32$), for a total of five types of receptive fields.
Multiscale features covering these small to large receptive fields are learned by adding them together.

In the original AdaptFormer, the number of dimensions is reduced once in the fully connected layer and then restored to the original number of dimensions in the fully connected layer to learn parameters with a low computational cost, and the same structure is used in SM-AdaptFormer.
Therefore, even when acquiring multi-scale features, a low dimensionality is an input, thus avoiding computational bloat.

\section{Experiments}
\label{sec:experiments}

\subsection{Datasets and Metrics}
\label{sub:dataset}

In the experiments, we assessed different types of image data from various domains with varying input image sizes: in-vehicle images, satellite images, microscopic images, endoscopic images, CT images, and transparent object images. 
We used two large datasets with more than 10,000 images and five smaller datasets with 1,000 images or less.
Specifically, we used the Cityscapes dataset \cite{cityscapes} (19 classes) for in-vehicle images and the Trans10k \cite{trans10k} dataset (3 classes) for transparent object images as large datasets.

As smaller datasets, we used the CamVid dataset \cite{camvid} (11 classes) for in-vehicle images, the Massachusetts Buildings dataset \cite{mbuilding} (M-Building, 2 classes) for satellite images, the ISBI2012 dataset \cite{isbi} (2 classes) for microscopic images, the Kvasir-SEG dataset \cite{kvasir} (2 classes) for endoscopic images, and the Synapse multi-organ dataset \cite{smo} (Synapse, 9 classes) for CT images. 
The pixel counts for each dataset are listed in \cref{tab:result-large} and \cref{tab:result-small}. 

In semantic segmentation, Intersection over Union (IoU), which indicates the overlap ratio between prediction and ground truth labels is generally used as an evaluation metric. 
Therefore, we used Mean IoU (mIoU) and the average IoU of all classes as the evaluation metrics.

\subsection{Training Conditions}
\label{sub:implementation}

In this paper, we used Pytorch library and trained the model using Adam optimizer for 200 epochs with a batch size of 8.
The learning rate was initially set to 0.005 and gradually decreased using the cosine learning rate scheduler \cite{cosinelr}.
For comparison, we used conventional CNN-based networks such as U-Net \cite{unet} and DeepLabv3+ \cite{deeplabv3plus}, as well as efficient fine-tuning methods using SAM: LoRA \cite{lora}, ConvLoRA \cite{convlora}, AdaptFormer \cite{adaptformer}, and SAMUS \cite{samus}.

For data pre-processing during training, we used random cropping, horizontal flipping, and random rotation for the CNN-based methods and GSAM. 
Other methods only accept images of fixed-size and therefore random cropping cannot be used.
Therefore, we only applied horizontal flipping and random rotation.
However, random rotation is not used for in-vehicle images and transparent object images.
This is because these two types of images have a clearly defined top and bottom, and do not require pre-processing by random rotation, which would change the top and bottom direction.
Image sizes for random cropping are listed in \cref{tab:result-large} and \cref{tab:result-small}.

\begin{table*}[t]
    \centering
    \caption{Comparison results on large datasets. Random cropping is applied only to U-Net, DeepLabv3+, and GSAM. The size of input images is $256\times256$ for SAMUS and $1024\times1024$ for the other comparison methods. 
    Among the proposed methods, the one that introduces only SM-AdaptFormer into the SAM is shown as SM-AdaptFormer, and the one that further supports variable input image sizes is shown as GSAM.
    The input image size of SM-AdaptFormer is $1024\times1024$.
    }
    \label{tab:result-large}
    \scalebox{0.9}{
    \begin{tabular*}{9.5cm}{@{\extracolsep{\fill}}rcc} \bhline{1.5pt}
    \multicolumn{1}{r}{Image type} & \multicolumn{1}{c}{Urban scene} & \multicolumn{1}{c}{Transparent object}\\
    \hline
    \multicolumn{1}{r}{Dataset} & \multicolumn{1}{c}{Cityscapes} & \multicolumn{1}{c}{Trans10k}\\
    \hline
    \multicolumn{1}{r}{Class} & \multicolumn{1}{c}{19} & \multicolumn{1}{c}{3}\\
    \hline
    \multicolumn{1}{r}{Image size} & \multicolumn{1}{c}{$1024 \times 2048$} & \multicolumn{1}{c}{Various}\\
    \hline 
    $\rm{Random\ crop}^ \ast$ & $512 \times 512$ & $256 \times 256$\\
    \hline
    U-Net \cite{unet}
    &67.30&72.18\\
    DeepLabv3+ \cite{deeplabv3plus}
    &67.04&70.82\\
    \hline 
    SAM \cite{sam}
    &57.15&83.37\\
    LoRA \cite{lora}
    &59.09&85.71\\
    ConvLoRA \cite{convlora}
    &62.43&86.47\\
    AdaptFormer \cite{adaptformer}
    &75.49&\textbf{\textcolor{red}{89.91}}\\
    SAMUS \cite{samus}
    &48.61&87.18\\
    \hline 
    SM-AdaptFormer
    &\textbf{\textcolor{red}{76.25}}&89.19\\
    GSAM
    &74.10&87.08\\
    \bhline{1.5pt}
    \end{tabular*}
    }
\end{table*}

\begin{table*}[t]
    \centering
    \caption{Comparison results on smaller datasets. Random cropping is applied only to U-Net, DeepLabv3+, and GSAM. The size of input images is $256\times256$ for SAMUS and $1024\times1024$ for the other comparison methods.}
    \label{tab:result-small}
    \scalebox{0.8}{
    \begin{tabular*}{15cm}{@{\extracolsep{\fill}}rccccc} \bhline{1.5pt}
    \multicolumn{1}{r}{Image type} & \multicolumn{1}{c}{Urban scene} & \multicolumn{1}{c}{Satellite} & \multicolumn{1}{c}{Biological}& \multicolumn{1}{c}{Endoscopic}& \multicolumn{1}{c}{CT}\\
    \hline
    \multicolumn{1}{r}{Dataset} & \multicolumn{1}{c}{CamVid} & \multicolumn{1}{c}{M-Building} & \multicolumn{1}{c}{ISBI2012}& \multicolumn{1}{c}{Kvasir-SEG}& \multicolumn{1}{c}{Synapse}\\
    \hline
    \multicolumn{1}{r}{Class} & \multicolumn{1}{c}{11} & \multicolumn{1}{c}{2} & \multicolumn{1}{c}{2} & \multicolumn{1}{c}{2} & \multicolumn{1}{c}{9}\\
    \hline
    \multicolumn{1}{r}{Image size} & \multicolumn{1}{c}{$360 \times 480$} & \multicolumn{1}{c}{$1500 \times 1500$} & \multicolumn{1}{c}{$256 \times 256$}& \multicolumn{1}{c}{$224 \times 224$}& \multicolumn{1}{c}{$224 \times 224$}\\
    \hline 
    \multirow{2}{*}{$\rm{Random\ crop}^ \ast$} & \multirow{2}{*}{$256 \times 256$} & \multirow{2}{*}{$512 \times 512$} & \multirow{2}{*}{$128 \times 128$}& $224 \times 224$& $224 \times 224$\\
    &&&&+ padding&+ padding\\
    \hline
    U-Net \cite{unet}
    &49.79&77.73&80.01&80.99&57.31\\
    DeepLabv3+ \cite{deeplabv3plus}
    &56.39&76.05&78.47
    &86.21&60.60\\
    \hline 
    SAM \cite{sam}
    &58.27&67.59&72.15&75.94&40.61\\
    LoRA \cite{lora}
    &65.20&76.76&79.18&82.20&39.08\\
    ConvLoRA \cite{convlora}
    &66.96&77.32&79.87&85.20&43.41\\
    AdaptFormer \cite{adaptformer}
    &\textbf{\textcolor{red}{74.80}}&80.46&80.46&88.53&61.28\\
    SAMUS \cite{samus}
    &48.42&49.87&78.64&88.28&20.66\\
    \hline 
    SM-AdaptFormer
    &73.99&80.41&80.48&\textbf{\textcolor{red}{88.76}}&66.06\\
    GSAM&67.21&\textbf{\textcolor{red}{80.69}}&\textbf{\textcolor{red}{80.53}}&87.83&\textbf{\textcolor{red}{72.78}}\\
    \bhline{1.5pt}
    \end{tabular*}
    }
\end{table*}

\subsection{Experimental Results}
\label{sub:experimantal-results}

\subsubsection{Quantitative Results.}

\cref{tab:result-large} and \cref{tab:result-small} show the quantitative results for each dataset. 
Regardless of the size of the dataset, GSAM achieved comparable or even higher accuracy than existing fine-tuning methods using SAM.
The red numbers in the table indicate the most accurate values.
Except for the Trans10k and the CamVid datasets, the proposed methods, SM-AdaptFormer and GSAM, showed the highest accuracy for the other five datasets.
Especially for the Synapse multi-organ dataset, which is CT images, SM-AdaptFormer and GSAM improved the accuracy by 4.78\% and 11.50\%, respectively, compared to AdaptFormer.
This result indicates that our proposed method may be highly effective in certain areas.
In addition, for all datasets, GSAM showed higher accuracy than the network composed of CNNs. 
This is considered to be due to the effectiveness of SAM itself, which is the underlying model based on Transformers, plus the learning of spatial information by the CNN-based SM-AdaptFormer and the effect of data expansion by random cropping, which took advantage of the benefits of each.

On the other hand, for the Trans10k and the CamVid datasets, the most accurate was AdaptFormer, while the second most accurate was SM-AdaptFormer which was the proposed method. 
The Trans10k dataset contains objects of relatively large size in the image.
In the case of such images, the SM-Adaptformer for extracting multi-scale information is considered to be less useful because the importance of information such as fine details is not high.
The CamVid dataset has many classes among the datasets, and it is considered more difficult to perform fine-tuning. 
The reason may be that the number of rates of Dilated Convolution set by SM-AdaptFormer was not appropriate because small objects were included.
However, the accuracy of SM-Adaptformer outperforms Adaptformer in the Cityscapes dataset, a dataset similar in systematics to the CamVid, which is considered that the optimal number of Dilated Convolution rates varies depending on the dataset.
The advantage of GSAM, however, is that it supports variable input image sizes.
The ability to input images with aspect ratios other than 1:1, such as the CamVid and Cityscapes datasets, in their original form without any smoothing or cropping is a unique advantage of GSAM among SAM fine-tuning methods.

Datasets such as Trans10k and Kvasir-SEG, which have relatively large objects in the images and fewer classes, are easier to perform fine-tuning on and show smaller differences in accuracy between methods.
For datasets such as these, the effectiveness of the GSAM advantage of performing random cropping is reduced, and the accuracy is not necessarily superior compared to other methods.
Although there may be a more appropriate size of the random cropping for GSAM, we can confirm that the accuracy of GSAM is significantly improved in comparison with that of CNN-based methods.
This can be attributed to the combination of multiscale spatial features using SM-AdaptFormer and the effectiveness of the SAM itself, which outperforms CNN-based models.

Based on the above results, it was confirmed that GSAM effectively acquires spatial features using SM-AdaptFormer and reduces computational costs compared to conventional SAM fine-tuning methods by supporting random cropping with variable-length inputs, while achieving equal or significantly higher accuracy.

\begin{figure*}[t]
\begin{center}
\includegraphics[scale=0.44]{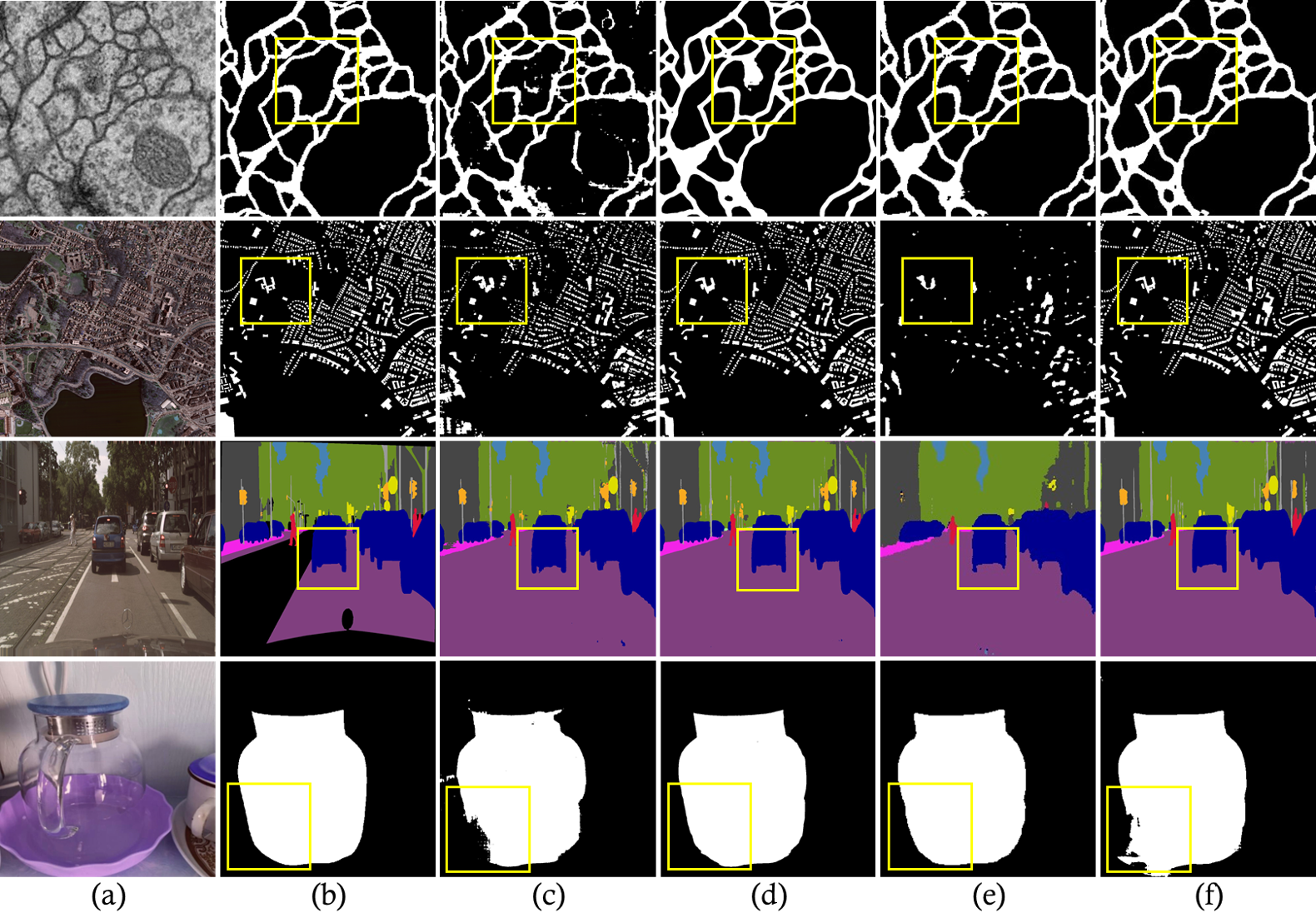}
\end{center}
\caption{Qualitative results. The first row is the results on the ISBI2012, the second is on the M-Building, the third is on the Cityscapes, and the fourth is on the Trans10k dataset. The Cityscapes dataset can be fed into GSAM with its original aspect ratio, but for simplicity of comparison, the same aspect ratios as other methods are shown. (a) Input image, (b) Ground truth, (c) SAM \cite{sam}, (d) AdaptFormer \cite{adaptformer}, (e) SAMUS \cite{samus}, (f) GSAM.}
\label{fig:kekka}
\end{figure*}

\subsubsection{Qualitative Results.}

\cref{fig:kekka} illustrates the qualitative results on four datasets. 
Specifically, the qualitative results are presented for the ISBI and the M-Building datasets, where GSAM had the highest accuracy among the comparison methods, and for the Cityscapes and the Trans10k datasets, where GSAM did not have the highest accuracy.

We first focus on the dataset in the top two rows.
These datasets contain the detailed structures and fine objects.
The results from the ISBI dataset show that GSAM reduces the over-detection of cell membrane classes compared to other SAM fine-tuning methods.
On the M-Building dataset, GSAM is able to segment small and complex shaped objects better than other methods.
These results show that the characteristics of GSAM, such as the ability to use random cropping and the ability to extract multi-scale features with SM-Adaptformer, are effective for datasets containing small objects and complex structures.

Next, we focus on the bottom two rows of the dataset.
These datasets have diverse classes or contain relatively large objects.
The Cityscapes dataset results show no particular advantage of GSAM over Adaptformer.
The Trans10k dataset results also show a significant failure in the segmentation of the lower part of the object when GSAM is used.
These factors may be because the number of rates of Dilated Convolution in SM-Adaptformer is always fixed and therefore not suitable for that data set.
In addition, GSAM, which aims to acquire multi-scale features, may not be suitable for datasets containing only large objects.

\begin{table*}[t]
    \centering
    \caption{Ablation Study of SM-Adaptformer on the ISBI2012 dataset. Comparative experiments were carried out by random cropping to $128 \times 128$ while maintaining the ability of GSAM to accept different input image sizes.}
    \scalebox{0.92}{
    \begin{tabular*}{6cm}{@{\extracolsep{\fill}}lccccc}\bhline{1.5pt}
    \multicolumn{1}{l}{Methods} & \multicolumn{1}{c}{mIoU} \\      
    \hline
        AdaptFormer \cite{adaptformer} &79.19 \\
        w/o ALL Convolutions &80.17 \\
        w/o ALL Dilated Convolutions &80.27 \\
        w/o $1 \times 1$ Convolution &80.25 \\
        w/o $3 \times 3$ Convolution &80.48 \\
        w/o Dilated Convolution(r=12) &80.32 \\
        w/o Dilated Convolution(r=24) &80.35 \\
        w/o Dilated Convolution(r=36) &80.42 \\
        SM-AdaptFormer(Ours) &\textcolor{red}{\textbf{80.53}} \\
    \bhline{1.5pt}
    \end{tabular*}
    \label{tab:ablation-sm}
    }
\end{table*}

\begin{table*}[t]
    \centering
    \caption{Comparison of MACs and segmentation accuracy for each method on the ISBI2012 dataset. Random cropping is only performed for GSAM, cropping to the number of pixels indicated by the "size". Input images with a size of $256\times256$ for SAMUS and $1024\times1024$ for the other comparison methods. 
    Finally, we adopted the experimental conditions that achieved the second highest segmentation accuracy with MACs of about 50\% of SAMUS. (shown in red)
    }
    \scalebox{0.92}{
    \begin{tabular*}{7cm}{@{\extracolsep{\fill}}lccc}\bhline{1.5pt}
    \multicolumn{1}{l}{Methods} & \multicolumn{1}{c}{MACs(G)} & \multicolumn{1}{c}{mIoU} \\      
    \hline
        SAM \cite{sam} &371.98&72.15 \\
        LoRA \cite{lora} &371.98&79.18 \\
        ConvLoRA \cite{convlora} &511.45&79.87 \\
        AdaptFormer \cite{adaptformer} &386.48&80.46 \\
        SAMUS \cite{samus} &145.87&78.64 \\
        \hline
        GSAM(size$=256\times256$) &270.33&\textbf{80.63} \\
        GSAM(size$=128\times128$) &\textcolor{red}{\textbf{74.07}}&\textcolor{red}{\textbf{80.53}} \\
        GSAM(size$=64\times64$) &18.53&78.76 \\
        GSAM(size$=32\times32$) &\textbf{7.42}&71.45 \\
    \bhline{1.5pt}
    \end{tabular*}
    \label{tab:ablation-macs}
    }
\end{table*}

\subsection{Ablation Study}
\label{sub:ablation}
\subsubsection{Effectiveness of SM-Adaptformer.}
We performed an ablation study on the SM-Adaptformer proposed as an internal module of GSAM. 
To test the effect of each component, we systematically removed each component from the SM-Adaptformer one by one. 
During this process, we maintained the ability of the GSAM to accept various input image sizes.
These results confirm that the convolutional layers of various scales in SM-Adaptformer significantly contribute to improving the segmentation accuracy. 
Additionally, the effectiveness of both standard and dilated convolutional layers was demonstrated.

From these findings, it is evident that SM-Adaptformer is more effective than AdaptFormer, which consists only of coupling layers and activation functions, due to its ability to acquire spatial information at multiple scales. 
However, in terms of extracting multi-scale features, it is particularly effective for datasets that include a wide range of objects from small objects to somewhat large objects, etc.

\begin{figure}[t]
\begin{center}
\includegraphics[scale=0.47]{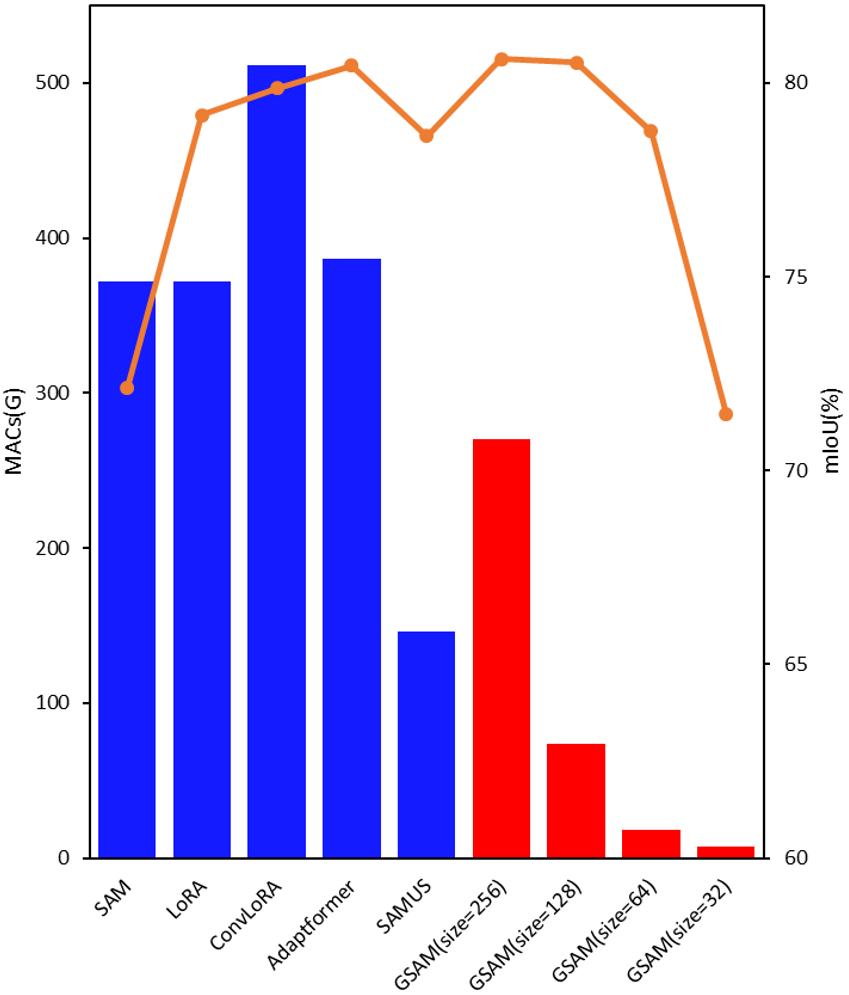}
\end{center}
\caption{The MACs and segmentation accuracy of each method on the ISBI2012 dataset are illustrated. The bar graphs are the MACs for each method and the line graphs are mIoU.}
\label{fig:macs}
\end{figure}

\subsubsection{Efficiency.}
\cref{tab:ablation-macs}, \cref{fig:macs} shows the results of our comparative experiments on the efficiency of GSAM.
As a comparison method, the MACs values of SAM, LoRA, ConvLoRA, AdaptFormer, and GSAM are compared when the size of the random cropping is changed. 
Since the number of input image sizes is fixed in conventional SAM fine-tuning methods, the computational cost becomes huge, and it can be seen that the computational cost is more than 300G MACs for all the methods except SAMUS. 
On the other hand, with GSAM, the computational cost decreases exponentially as the size of the random crop is reduced.
In particular, when the input image size is $128 \times 128$, the segmentation accuracy outperforms all conventional methods, despite a computational cost of around half that of SAMUS.
The SM-Adaptformer included in GSAM has a relatively complex structure to acquire multi-scale features and improve segmentation accuracy, but by using PEG and CNN to support variable-size input, the input image size can be reduced, which reduces the computational cost has been significantly reduced.
Based on these results, we expect GSAM to be widely used as an efficient approach for fine-tuning SAM in the future, as it significantly reduces the computational cost and allows for highly accurate segmentation.

\section{Conclusion}
\label{sec:conclusion}
In this paper, a novel fine-tuning method of SAM, GSAM, is proposed to handle variable input image sizes for SAM. 
GSAM is the first method to allow random cropping for SAM during training and significantly reduce the computational cost during training. 
From evaluation experiments on datasets with various input image sizes, we have confirmed that GSAM can train more efficiently than the conventional fine-tuning methods for SAM and can achieve the same or better segmentation accuracy.

In the future, we would like to address the problems caused by the use of Dilated Convolution with fixed rate values within SM-Adaptformer, and achieve the associated increase in versatility.
Since GSAM trains all the weight parameters of Decoder of SAM, we are considering adding the LoRA structure to Decoder to train Decoder more efficiently.


%
%
\bibliographystyle{splncs04}
\bibliography{main}

@String(ECCV  = {Eur. Conf. Comput. Vis.})

@String(ECCV  = {ECCV})

@article{imagenet,
  title={Imagenet classification with deep convolutional neural networks},
  author={Krizhevsky, Alex and Sutskever, Ilya and Hinton, Geoffrey E},
  journal={Advances in neural information processing systems},
  volume={25},
  year={2012}
}

@inproceedings{resnet,
  title={Deep residual learning for image recognition},
  author={He, Kaiming and Zhang, Xiangyu and Ren, Shaoqing and Sun, Jian},
  booktitle={Proceedings of the IEEE conference on computer vision and pattern recognition},
  pages={770--778},
  year={2016}
}

@article{bert,
  title={Bert: Pre-training of deep bidirectional transformers for language understanding},
  author={Devlin, Jacob and Chang, Ming-Wei and Lee, Kenton and Toutanova, Kristina},
  journal={arXiv preprint arXiv:1810.04805},
  year={2018}
}

@inproceedings{clip,
  title={Learning transferable visual models from natural language supervision},
  author={Radford, Alec and Kim, Jong Wook and Hallacy, Chris and Ramesh, Aditya and Goh, Gabriel and Agarwal, Sandhini and Sastry, Girish and Askell, Amanda and Mishkin, Pamela and Clark, Jack and others},
  booktitle={International conference on machine learning},
  pages={8748--8763},
  year={2021},
  organization={PMLR}
}

@inproceedings{sam,
  title={Segment anything},
  author={Kirillov, Alexander and Mintun, Eric and Ravi, Nikhila and Mao, Hanzi and Rolland, Chloe and Gustafson, Laura and Xiao, Tete and Whitehead, Spencer and Berg, Alexander C and Lo, Wan-Yen and others},
  booktitle={Proceedings of the IEEE/CVF International Conference on Computer Vision},
  pages={4015--4026},
  year={2023}
}

@article{lora,
  title={Lora: Low-rank adaptation of large language models},
  author={Hu, Edward J and Shen, Yelong and Wallis, Phillip and Allen-Zhu, Zeyuan and Li, Yuanzhi and Wang, Shean and Wang, Lu and Chen, Weizhu},
  journal={arXiv preprint arXiv:2106.09685},
  year={2021}
}

@article{adaptformer,
  title={Adaptformer: Adapting vision transformers for scalable visual recognition},
  author={Chen, Shoufa and Ge, Chongjian and Tong, Zhan and Wang, Jiangliu and Song, Yibing and Wang, Jue and Luo, Ping},
  journal={Advances in Neural Information Processing Systems},
  volume={35},
  pages={16664--16678},
  year={2022}
}

@article{samus,
  title={Samus: Adapting segment anything model for clinically-friendly and generalizable ultrasound image segmentation},
  author={Lin, Xian and Xiang, Yangyang and Zhang, Li and Yang, Xin and Yan, Zengqiang and Yu, Li},
  journal={arXiv preprint arXiv:2309.06824},
  year={2023}
}

@inproceedings{unet,
  title={U-net: Convolutional networks for biomedical image segmentation},
  author={Ronneberger, Olaf and Fischer, Philipp and Brox, Thomas},
  booktitle={Medical image computing and computer-assisted intervention--MICCAI 2015: 18th international conference, Munich, Germany, October 5-9, 2015, proceedings, part III 18},
  pages={234--241},
  year={2015},
  organization={Springer}
}

@inproceedings{deeplabv3plus,
  title={Encoder-decoder with atrous separable convolution for semantic image segmentation},
  author={Chen, Liang-Chieh and Zhu, Yukun and Papandreou, George and Schroff, Florian and Adam, Hartwig},
  booktitle={Proceedings of the European conference on computer vision (ECCV)},
  pages={801--818},
  year={2018}
}

@article{peg,
  title={Conditional positional encodings for vision transformers},
  author={Chu, Xiangxiang and Tian, Zhi and Zhang, Bo and Wang, Xinlong and Shen, Chunhua},
  journal={arXiv preprint arXiv:2102.10882},
  year={2021}
}

@article{dilatedconv,
  title={Multi-scale context aggregation by dilated convolutions},
  author={Yu, Fisher and Koltun, Vladlen},
  journal={arXiv preprint arXiv:1511.07122},
  year={2015}
}

@article{convlora,
  title={Convolution Meets LoRA: Parameter Efficient Finetuning for Segment Anything Model},
  author={Zhong, Zihan and Tang, Zhiqiang and He, Tong and Fang, Haoyang and Yuan, Chun},
  journal={arXiv preprint arXiv:2401.17868},
  year={2024}
}

@article{samed,
  title={Customized segment anything model for medical image segmentation},
  author={Zhang, Kaidong and Liu, Dong},
  journal={arXiv preprint arXiv:2304.13785},
  year={2023}
}

@inproceedings{trans10k,
  title={Segmenting transparent objects in the wild},
  author={Xie, Enze and Wang, Wenjia and Wang, Wenhai and Ding, Mingyu and Shen, Chunhua and Luo, Ping},
  booktitle={Computer Vision--ECCV 2020: 16th European Conference, Glasgow, UK, August 23--28, 2020, Proceedings, Part XIII 16},
  pages={696--711},
  year={2020},
  organization={Springer}
}

@article{camvid,
  title={Semantic object classes in video: A high-definition ground truth database},
  author={Brostow, Gabriel J and Fauqueur, Julien and Cipolla, Roberto},
  journal={Pattern recognition letters},
  volume={30},
  number={2},
  pages={88--97},
  year={2009},
  publisher={Elsevier}
}

@inproceedings{cityscapes,
  title={The cityscapes dataset for semantic urban scene understanding},
  author={Cordts, Marius and Omran, Mohamed and Ramos, Sebastian and Rehfeld, Timo and Enzweiler, Markus and Benenson, Rodrigo and Franke, Uwe and Roth, Stefan and Schiele, Bernt},
  booktitle={Proceedings of the IEEE conference on computer vision and pattern recognition},
  pages={3213--3223},
  year={2016}
}

@phdthesis{mbuilding,
    author = {Volodymyr Mnih},
    title = {Machine Learning for Aerial Image Labeling},
    school = {University of Toronto},
    year = {2013}
}

@misc{isbi,
key = {Howard Hughes Medical Institute},
title  = {Segmentation of neuronal structures in EM stacks challenge},
year = "2012",
howpublished  = {\url{https://imagej.net/events/isbi-2012-segmentation-challenge}},
}

@inproceedings{kvasir,
  title={Kvasir-seg: A segmented polyp dataset},
  author={Jha, Debesh and Smedsrud, Pia H and Riegler, Michael A and Halvorsen, P{\aa}l and De Lange, Thomas and Johansen, Dag and Johansen, H{\aa}vard D},
  booktitle={MultiMedia Modeling: 26th International Conference, MMM 2020, Daejeon, South Korea, January 5--8, 2020, Proceedings, Part II 26},
  pages={451--462},
  year={2020},
  organization={Springer}
}

@misc{smo,
key = {Sage Bionetworks},
title  = {Multi-Atlas Labeling Beyond the Cranial Vault - Workshop and Challenge},
year = "2015",
howpublished  = {\url{https://www.synapse.org/#!Synapse:syn3193805/wiki/217789}},
}

@article{cosinelr,
  title={Sgdr: Stochastic gradient descent with warm restarts},
  author={Loshchilov, Ilya and Hutter, Frank},
  journal={arXiv preprint arXiv:1608.03983},
  year={2016}
}

@article{vit,
  author       = {Alexey Dosovitskiy and
                  Lucas Beyer and
                  Alexander Kolesnikov and
                  Dirk Weissenborn and
                  Xiaohua Zhai and
                  Thomas Unterthiner and
                  Mostafa Dehghani and
                  Matthias Minderer and
                  Georg Heigold and
                  Sylvain Gelly and
                  Jakob Uszkoreit and
                  Neil Houlsby},
  title        = {An Image is Worth 16x16 Words: Transformers for Image Recognition
                  at Scale},
  journal      = {CoRR},
  volume       = {abs/2010.11929},
  year         = {2020},
  url          = {https://arxiv.org/abs/2010.11929},
  eprinttype    = {arXiv},
  eprint       = {2010.11929},
  bibsource    = {dblp computer science bibliography, https://dblp.org}
}

@article{gpt4,
  title={Sparks of artificial general intelligence: Early experiments with gpt-4},
  author={Bubeck, S{\'e}bastien and Chandrasekaran, Varun and Eldan, Ronen and Gehrke, Johannes and Horvitz, Eric and Kamar, Ece and Lee, Peter and Lee, Yin Tat and Li, Yuanzhi and Lundberg, Scott and others},
  journal={arXiv preprint arXiv:2303.12712},
  year={2023}
}

@article{llama,
  title={Llama: Open and efficient foundation language models},
  author={Touvron, Hugo and Lavril, Thibaut and Izacard, Gautier and Martinet, Xavier and Lachaux, Marie-Anne and Lacroix, Timoth{\'e}e and Rozi{\`e}re, Baptiste and Goyal, Naman and Hambro, Eric and Azhar, Faisal and others},
  journal={arXiv preprint arXiv:2302.13971},
  year={2023}
}

@article{transformer,
  title={Attention is all you need},
  author={Vaswani, Ashish and Shazeer, Noam and Parmar, Niki and Uszkoreit, Jakob and Jones, Llion and Gomez, Aidan N and Kaiser, {\L}ukasz and Polosukhin, Illia},
  journal={Advances in neural information processing systems},
  volume={30},
  year={2017}
}

@inproceedings{sd,
  title={High-resolution image synthesis with latent diffusion models},
  author={Rombach, Robin and Blattmann, Andreas and Lorenz, Dominik and Esser, Patrick and Ommer, Bj{\"o}rn},
  booktitle={Proceedings of the IEEE/CVF conference on computer vision and pattern recognition},
  pages={10684--10695},
  year={2022}
}

@inproceedings{unet++,
  title={Unet++: A nested u-net architecture for medical image segmentation},
  author={Zhou, Zongwei and Rahman Siddiquee, Md Mahfuzur and Tajbakhsh, Nima and Liang, Jianming},
  booktitle={Deep Learning in Medical Image Analysis and Multimodal Learning for Clinical Decision Support: 4th International Workshop, DLMIA 2018, and 8th International Workshop, ML-CDS 2018, Held in Conjunction with MICCAI 2018, Granada, Spain, September 20, 2018, Proceedings 4},
  pages={3--11},
  year={2018},
  organization={Springer}
}

@article{r2unet,
  title={Recurrent residual convolutional neural network based on u-net (r2u-net) for medical image segmentation},
  author={Alom, Md Zahangir and Hasan, Mahmudul and Yakopcic, Chris and Taha, Tarek M and Asari, Vijayan K},
  journal={arXiv preprint arXiv:1802.06955},
  year={2018}
}

@inproceedings{pspnet,
  title={Pyramid scene parsing network},
  author={Zhao, Hengshuang and Shi, Jianping and Qi, Xiaojuan and Wang, Xiaogang and Jia, Jiaya},
  booktitle={Proceedings of the IEEE conference on computer vision and pattern recognition},
  pages={2881--2890},
  year={2017}
}

@article{deeplab,
  title={Deeplab: Semantic image segmentation with deep convolutional nets, atrous convolution, and fully connected crfs},
  author={Chen, Liang-Chieh and Papandreou, George and Kokkinos, Iasonas and Murphy, Kevin and Yuille, Alan L},
  journal={IEEE transactions on pattern analysis and machine intelligence},
  volume={40},
  number={4},
  pages={834--848},
  year={2017},
  publisher={IEEE}
}

@article{deeplabv3,
  title={Rethinking atrous convolution for semantic image segmentation},
  author={Chen, Liang-Chieh and Papandreou, George and Schroff, Florian and Adam, Hartwig},
  journal={arXiv preprint arXiv:1706.05587},
  year={2017}
}

@article{transunet,
  title={Transunet: Transformers make strong encoders for medical image segmentation},
  author={Chen, Jieneng and Lu, Yongyi and Yu, Qihang and Luo, Xiangde and Adeli, Ehsan and Wang, Yan and Lu, Le and Yuille, Alan L and Zhou, Yuyin},
  journal={arXiv preprint arXiv:2102.04306},
  year={2021}
}

@inproceedings{swinunet,
  title={Swin-unet: Unet-like pure transformer for medical image segmentation},
  author={Cao, Hu and Wang, Yueyue and Chen, Joy and Jiang, Dongsheng and Zhang, Xiaopeng and Tian, Qi and Wang, Manning},
  booktitle={European conference on computer vision},
  pages={205--218},
  year={2022},
  organization={Springer}
}
\end{document}